\documentclass[letterpaper]{article} % DO NOT CHANGE THIS
\usepackage{aaai23}  % DO NOT CHANGE THIS
\usepackage{times}  % DO NOT CHANGE THIS
\usepackage{helvet}  % DO NOT CHANGE THIS
\usepackage{courier}  % DO NOT CHANGE THIS
\usepackage[hyphens]{url}  % DO NOT CHANGE THIS
\usepackage{graphicx} % DO NOT CHANGE THIS
\usepackage{natbib}  % DO NOT CHANGE THIS AND DO NOT ADD ANY OPTIONS TO IT
\usepackage{caption} % DO NOT CHANGE THIS AND DO NOT ADD ANY OPTIONS TO IT
\usepackage{algorithm}
\usepackage{algorithmic}

\usepackage{newfloat}
\usepackage{listings}
\DeclareCaptionStyle{ruled}{labelfont=normalfont,labelsep=colon,strut=off} % DO NOT CHANGE THIS
\floatstyle{ruled}
\newfloat{listing}{tb}{lst}{}
\floatname{listing}{Listing}
\usepackage{tabularx}
\usepackage{algorithm}
\usepackage{algorithmic}
\usepackage{subfigure}
\usepackage{graphicx}
\usepackage{easyReview}
\usepackage{multirow}
\usepackage{xspace}
\usepackage{balance}
\usepackage{amsfonts, amsmath, amssymb}
\usepackage{url}
\usepackage{easyReview}

\newcommand{\bS}{\mathbf{S}}
\newcommand{\by}{\mathbf{y}}
\newcommand{\bP}{\mathbf{P}}

\title{SLOTH: Structured Learning and Task-based Optimization for Time Series Forecasting on Hierarchies}
\author{
Fan Zhou, Chen Pan, Lintao Ma, Yu Liu, Shiyu Wang, James Zhang, Xinxin Zhu,  Xuanwei Hu, Yunhua Hu, Yangfei Zheng, Lei Lei, Yun Hu
}
\affiliations{
    Ant Group\\
    \{hanlian.zf, bopu.pc, lintao.mlt, nuoman.ly, weiming.wsy, james.z, qinrui.zxx, xuanwei.hxw, yangfei.zyf, jason.ll, huyun.h\}@antgroup.com, wugou.hyh@alibaba-inc.com
}

\usepackage{bibentry}
\begin{document}

\maketitle

\begin{abstract}
Multivariate time series forecasting with hierarchical structure is widely used in real-world applications,
e.g., sales predictions for the geographical hierarchy formed by cities, states, and countries.
The \textit{hierarchical time series} (HTS) forecasting includes two sub-tasks, i.e., \textit{forecasting} and \textit{reconciliation}.  In the previous works, hierarchical information is only integrated in the reconciliation step to maintain coherency, but not in forecasting step for accuracy improvement.
% neglecting the forecasting step that could help further improve accuracy.
In this paper, we propose two novel tree-based feature integration mechanisms, i.e., \textit{top-down convolution} and \textit{bottom-up attention} to leverage the information of the hierarchical structure to improve the forecasting performance.  
Moreover,  unlike most previous reconciliation methods which either rely on strong assumptions or focus on coherent constraints only, we utilize deep neural optimization networks, which not only achieve coherency without any assumptions, but also allow more flexible and realistic constraints to achieve task-based targets, e.g., lower under-estimation penalty and meaningful decision-making loss to facilitate the subsequent downstream tasks.
Experiments on real-world datasets demonstrate that our tree-based feature integration mechanism achieves superior performances on hierarchical forecasting tasks compared to the state-of-the-art methods, and our neural optimization networks can be applied to real-world tasks effectively without any additional effort under coherence and task-based constraints.
\end{abstract}

\section{Introduction}
Forecasting for multiple time series with complex hierarchical structure has been applied in various real-world problems \cite{dangerfield1992top, athanasopoulos2009hierarchical, liu2018flexible, jeon2018reconciliation}. For instance, the international sales forecasting of transnational corporations needs to consider geographical hierarchies involving the levels of city, state, and country \cite{han2021simultaneously}. Moreover, retail sales can also be divided into different groups according to the product category
to form another hierarchy.
The combination of different hierarchies can form a more complicated but realistic virtual topology, e.g., the geographical hierarchies (of different regions) and the commodity hierarchies (of varies categories) are nested to be parts of the supply chain
in the retail sector.

The critical challenge in hierarchical forecasting tasks lies in producing accurate prediction results while satisfying aggregation (coherence) constraints \cite{ben2019regularized}.
Specifically, the hierarchical structure in these time series implies coherence constraint, i.e., time series at upper levels are the aggregation of those at lower levels. However,  independent forecasts from a prediction model (called {\it base forecasts}) are unlikely to satisfy coherence constraint. 

Previous work on hierarchical forecasting mainly focuses on a procedure of two separate stages~\cite{hyndman2016fast, ben2019regularized, corani2020probabilistic, anderer2021forecasting}: In the first stage, {\it base forecasts} are generated independently for each time series in the hierarchy; in the second stage, these forecasts are adjusted via \textit{reconciliation} to derive coherent results.  
The base forecasts are obtained by univariate or multivariate time series models, such as traditional forecasting methods (e.g., ARIMA~\cite{ben2019regularized}) and deep learning techniques (e.g. DeepVar~\cite{salinas2019high}).  However,  both approaches ignore the information of hierarchical structure for prediction.
As for reconciliation, traditional statistical methods mostly rely on strong assumption, such as unbiased forecasts and Gaussian noises (e.g., MinT~\cite{wickramasuriya2019optimal}), which are often inconsistent with  non-Gaussian/non-linear real-world \textit{hierarchical time series} (HTS) data. A notable exception is the approach proposed in \cite{rangapuram2021end}, promising coherence by a closed-form projection to achieve an end-to-end reconciliation without any assumptions. 
However, this method may introduce huge adjustments on the original forecasts for coherency, making the final forecasts unreasonable.
Moreover, even coherent forecasters might still be impractical for some real-world tasks with further practical operational or task-related constraints, such as inventory management and resource scheduling.  
These reconciliation concerns can be addressed by imposing more realistic constraints to control the scale and optimize the task-based targets for the downstream tasks, which can be achieved by \textit{deep neural optimization layer} (OptNet) \cite{amos2017optnet}. 

In this work, we provides an end-to-end framework to generate coherent forecasts of hierarchical time series that incorporates the hierarchical structure in the prediction process, while taking task-based constraints and targets into consideration. In detail, our contributions to HTS forecasting and aligned decision-making can be summarised as follows:
\begin{itemize}
\item We propose two tree-based mechanisms, including top-down convolution and bottom-up attention, to leverage hierarchical structure information (through feature fusion of all levels in the hierarchy) for performance improvement of the base forecasts. To the best of our knowledge, our approach is the first model that harnesses the power of deep learning to exploit the complicated structure information of HTS.
\item We provide a flexible end-to-end learning framework to unify the goals of the forecasting and decision-making by employing a deep differentiable convex optimization layer (OptNet), which not only achieves controllable reconciliation 
without any assumptions, but also adapts to more practical task-related constraints and targets to solve the real-world 
problems without any explicit post-processing step.

\item Extensive experiments on real-world hierarchical datasets from various industrial domains demonstrate that our proposed approach achieves significant improvements over the state-of-the-art baseline methods, and our approach has been deployed online to cloud resource scheduling project in Ant Group.
\end{itemize}

\section{Preliminaries}
\label{section: preliminaries}
\begin{figure}[h]
  \centering
  \includegraphics[width=0.5\linewidth]{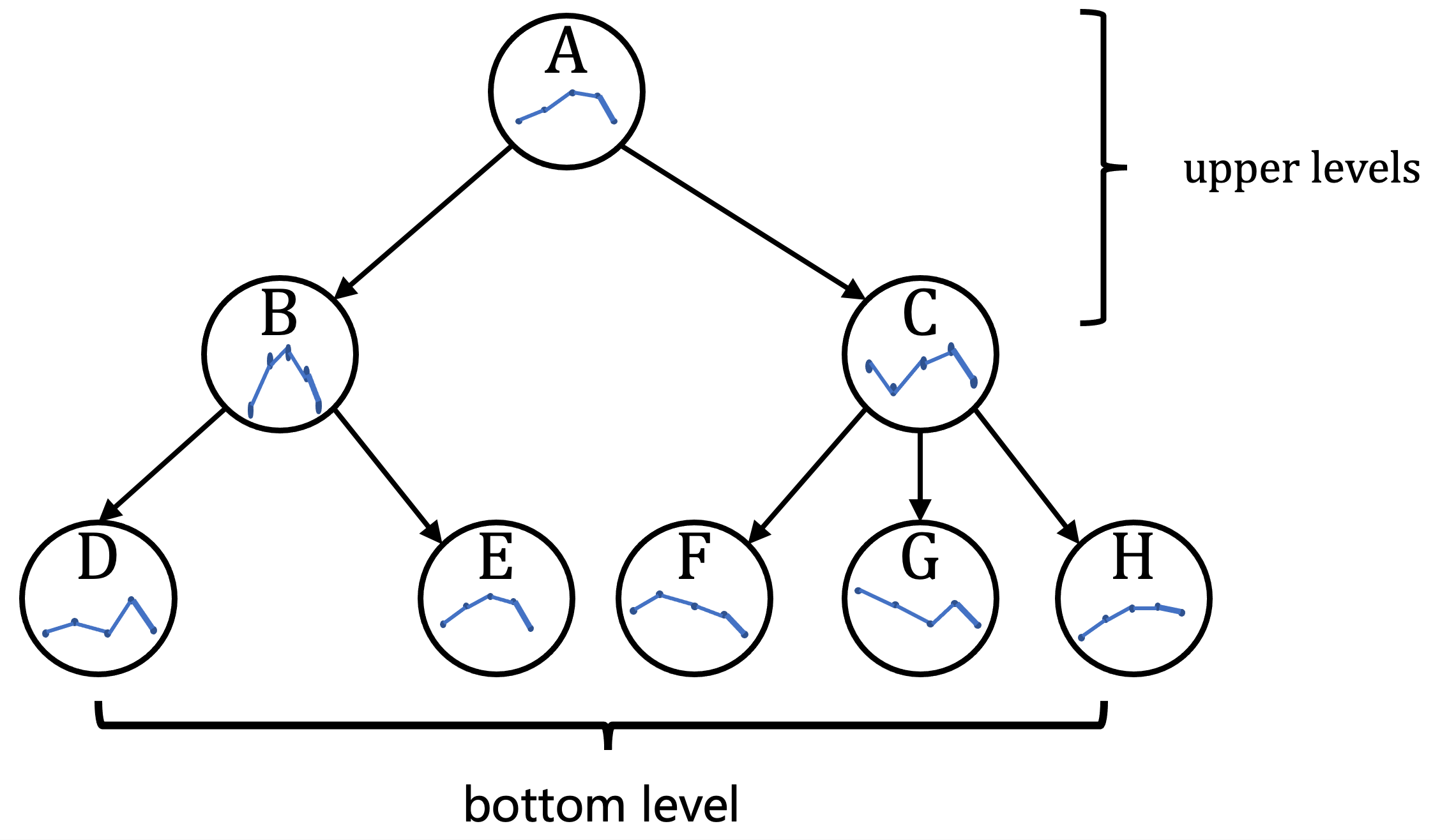}
  \caption{An example of HTS structure for $n=8$ time series with $m=5$ bottom-level series and $r=3$ upper-level series.}
  \label{fig:hier_timeseries}
\end{figure}

A hierarchical time series can be denoted as a tree structure  (see Figure \ref{fig:hier_timeseries}) with linear aggregation constraints, expressed by aggregation matrix $\mathbf{S}\in \mathbb{R}^{n \times m}$ ($m$ is the number of bottom-level nodes, and $n$ is the total number of nodes).  In the hierarchy, each node represents a time series, to be predicted over a time horizon. 

Given a time horizon $t\in \{1,2,\dots,T\}$, we use $y_{i,t}\in \mathbb{R}$ to denote the values of the $i$-th component of a multivariate hierarchical time series, where $i\in \{1,2,\dots,n\}$ is the index of the individual univariate time series.  Here we assume that the index $i$ abides by the level-order traversal of the hierarchical tree, going from left to right at each level. 
% {\color{red} (remove them?), and we use $x^i_{t}$ to denote the covariate vector of univariate time series $i$ at time step $t$.} 

In the tree structure, the time series of leaf nodes are called the \textit{bottom-level} series $\mathbf{b}_{t}\in \mathbb{R}^{m}$, and those of the remaining nodes are termed \textit{upper-levels} series $\mathbf{u}_{t}\in \mathbb{R}^{r}$. Obviously, the total number of nodes $n=r+m$, and $\by_{t} :=[\mathbf{u}_{t},\mathbf{b}_{t}]^{\mathsf{T}} \in  \mathbb{R}^{n} $ contains observations at time $t$ for all levels, which satisfies  
\begin{equation}
\label{eq_sum}
\by_{t}=\bS \mathbf{b}_{t},
\end{equation}
where $\bS\in \{0,1\}^{n \times m}$ is an aggregation matrix. 

Taking the HTS in Figure~\ref{fig:hier_timeseries} as an example, the aggregation matrix 
% \replace{takes}
{is in} the form:

\begin{equation*} %\label{eq:hts_matrix}
    \bS = 
    \begin{bmatrix}
        \bS_{\text{sum}}\\
        \mathbf{I}_5
    \end{bmatrix}
    \begin{bmatrix}
    \begin{array}{ccccc}
        1 & 1 & 1 &1 &1 \\
        1 & 1 & 0 &0 &0 \\
        0 & 0 & 1 &1 &1 \\
        % \hline \\
        \hdotsfor{5}\\
        % & & & & \\
        & & \mathbf{I}_5 & &\\
        % & & & &
     \end{array} 
     \end{bmatrix}
     ,
\end{equation*}
where $\mathbf{I}_5$ is an identity matrix of size 5. The total number of series in the hierarchy is $n=3+5$. At each time step $t$, 
\begin{equation*}
    \begin{aligned}
        \mathbf{u}_{t}&=[y_{A,t},y_{B,t},y_{C,t}] \in \mathbb{R}^{3}, \\
        \mathbf{b}_{t}&=[y_{D,t},y_{E,t},y_{F,t},y_{G,t},y_{H,t}] \in \mathbb{R}^{5},\\
        y_{A,t}&=y_{B,t}+y_{C,t}=y_{D,t}+y_{E,t}+y_{F,t}+y_{G,t}+y_{H,t}.
    \end{aligned}
\end{equation*}

The coherence constraint of Eq.~\eqref{eq_sum} can be represented as ~\cite{rangapuram2021end}
\begin{equation}
\label{eq:coherent_co}
\mathbf{A}\by_{t}=\mathbf{0},
\end{equation}
where $\mathbf{A}:=[\mathbf{I}_{r}|-\bS_{\text{sum}}] \in \{0,1\}^{r \times n}$, $\mathbf{0}$ is an $r$-vector of zeros, and $\mathbf{I}_{r}$ is a $r \times r $ identity matrix.

\begin{figure*}[h]
  \centering
  \includegraphics[width=0.85\textwidth]{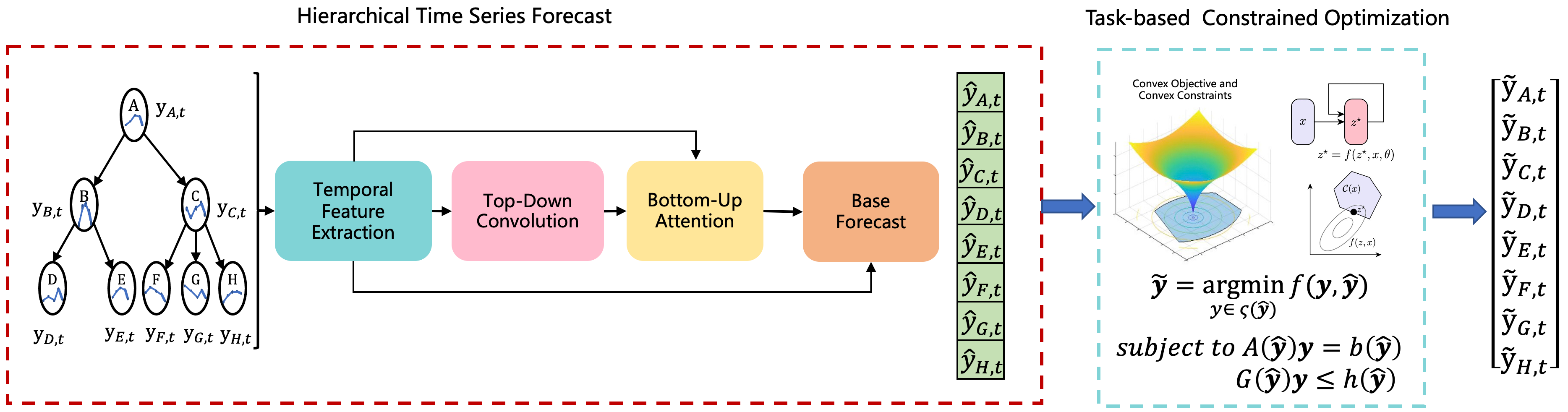}
  \caption{The Architecture of SLOTH: the red dashed box is the hierarchical forecasting component including the temporal feature
extraction module, top-down convolution module, bottom-up attention module, and base forecasting module. The base forecasting module
generates predictions without reconciliation. The light blue dashed box is the task-based optimization module that
generates the reconciliation forecasts and achieves task-based targets for real-world scenarios.}
  \label{fig_arch}
\end{figure*}

\section{Method}
In this section, we introduce our framework (SLOTH), which not only can improve prediction performance by integrating hierarchical structure into forecasting, but also achieves task-based goals with deep neural optimization layer fulfilling coherence constraint. Figure~\ref{fig_arch} illustrates the architecture of SLOTH, consisting of two main components: 
\begin{itemize}
    \item a \textit{structured hierarchical forecasting module} that produces forecasts over the prediction horizon across all nodes, utilizing both top-down convolution and bottom-up attention to enhance the {temporal} features for each node.
    \item a \textit{task-based constrained optimization module} that leverages OptNet to satisfy the coherence constraint and provide a flexible module for real-world tasks with more complex practical constraints and targets.
\end{itemize}

\subsection{Structured Hierarchical Forecasting}
In this section, we introduce the  \textit{structured hierarchical learning module}, which integrates dynamic features from the hierarchical structure to produce better base forecasts.

\subsubsection{\textbf{Temporal Feature Extraction Module}}
This module extracts temporal features of each node as follows:
% \begin{equation}
\begin{align}
\label{eq:uni}
    \bar{h}^i_t &= \text{UPDATE}(\bar{h}^i_{t-1}, x^i_t; \mathbf{\theta}), \; i \in \{{1,2, \dots, n}\},\\
% \end{equation}
% \begin{equation*}
    \bar{\mathbf{H}}_t & = [{\bar{h}^1_t, \bar{h}^2_t,  \dots, \bar{h}^{n}_t}],\notag
% \end{equation*}
\end{align}
where ${x}^i_t$ is the covariates of node $i$ at time $t$, $\bar{h}_{t-1}^i$ is the hidden feature of the previous time step $t-1$, $\mathbf{\theta}$ is the model parameters shared across all nodes, and $\textit{UPDATE}(\cdot)$ is the pattern extraction function. 
Any recurrent-type neural network can be adopted as the temporal feature extraction module, such as the RNN variants \cite{hochreiter1997long, chung2014empirical}, TCN \cite{bai2018empirical}, WAVENET~\cite{van2016wavenet}, and NBEATS \cite{oreshkin2019n}.  We use GRU ~\cite{chung2014empirical} in our experiments due to its simplicity.

It is worth noting that we process each node independently in this module, which is in contrast to some existing works that utilize multivariate model to extract the relationship between time series~\cite{rangapuram2021end}. It is unnecessary for our framework because we leverage the hierarchical structure in feature integration step, which we believe is enough to characterize the relationship between nodes. We reduce the time complexity from O($n^2$) to O($n$) accordingly.

\subsubsection{\textbf{Top-Down Convolution Module (TD-Conv)}}
\label{section:top_down}
This module incorporates structural information to dynamic pattern by integrating temporal features (e.g., trends and seasonality) from nodes at the top level into those at the bottom level to enhance the temporal stability. In other words, clearer seasonality and more smooth evolving \cite{taieb2021hierarchical}.

Most of the previous methods indicate that time series of nodes at the upper level are easier to predict than nodes at the bottom level, which implies the dynamic pattern at the top level is more stable. 
Therefore, the bottom nodes can use their ancestors' features at top levels to improve prediction performance.
Similar to Tree Convolution \cite{mou2016convolutional}, our approach introduces a top-down convolution mechanism by extracting effective top-level temporal patterns to denoise the feature of bottom nodes and increase stability.

\begin{figure}[htbp]
  \centering
  \includegraphics[width=0.7\linewidth]{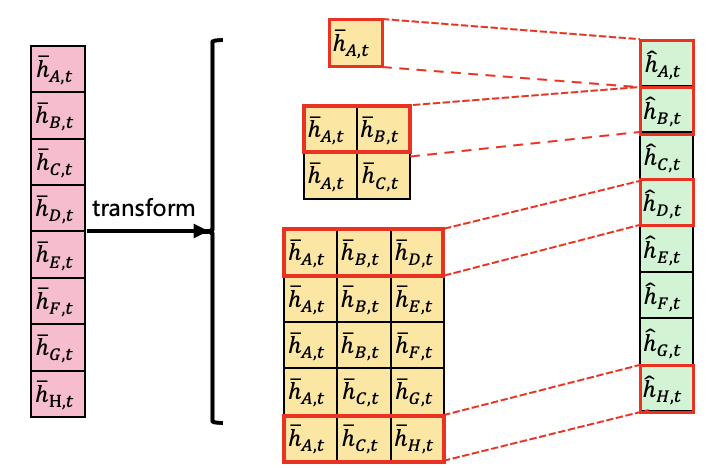}
  \caption{Top-Down Convolution Architecture. Given the temporal features of each node (pink boxes) from the temporal feature extraction module, our approach transforms these hidden features into matrices (yellow boxes) for fast convolution. The right part with red dashed lines demonstrates the integration of ancestors' features with CNN, and the green boxes represent the integration feature.}
  \label{fig:top_down}
\end{figure}

The top-down convolution mechanism shown in Figure~\ref{fig:top_down} is based on the outputs of the univariate forecasting model. We want to highlight that all nodes share the forecasting model to obtain the temporal features. 
% \replace{Please note that the value of each bottom node is appended to all its ancestors' values.}
{Please note that ancestor's value are the sum of all children nodes. }  In other words, ancestors' features are helpful to predict the considered node, which calls for the integration of the features of both levels for better prediction.

In order to reduce the computational complexity, the hidden states are reorganized into matrices (yellow boxes in the middle part of Figure~\ref{fig:top_down}) after temporal hidden features are obtained. Each row of the matrices represents the feature concatenation of nodes from considered node to root in the hierarchy.
We then apply convolution neural networks (CNNs) to each row of those matrices to aggregate the temporal patterns, since it is well known that CNN is an effective tool to integrate valid information from hidden features.

The computation complexity is very high if the convolution is applied on $\bar{\mathbf{H}}_t$ because it is not sorted as tree structure index. To accelerate the computation, we  transform the  hierarchical structure to a series of matrix forms to speed up the convolution process as shown in the middle part of
Figure~\ref{fig:top_down}. Specifically, the nodes' hidden features $\mathbf{\bar{H}}_t$ in the shape of $(n, d_h)$ are reorganized into a matrix form $\mathbf{\bar{H}}^{\prime}_t = \{\bar{h'}^{1}_t,..., \bar{h'}^{n}_t\}$, where $\bar{h'}^i_t = [\bar{h}^1_t, \dots, \bar{h}^{i}_t]$ is the concatenation of temporal features $\bar{h}_t$ of all the ancestors of node $i$ in the shape of $(l_i, d_h)$, where $n$ is number of nodes, $l_i$ is the number of levels of node $i$, and $d_h$ is the dimension. Please note that $\mathbf{\bar{H}}^{\prime}_t $ is not an actual matrix but a union of matrices of different dimensions, where the first dimension of each $\bar{h'}_t$ is the level index.
Then we apply convolution to $\bar{h'}_t^i$ to integrate temporal features of all ancestors of $i$ as follows
\begin{align}
    \label{eq:conv}
    \hat{h}_t^i &= \text{Conv}_{l_i}(\bar{h'}^i_t; \Theta_{l_i}) = \sum_{k=1}^{l_i} w_k  \bar{h}^{l_i -k}_t,\\
     \mathbf{\hat{H}}_t &= [{\hat{h}^1_t, \hat{h}^2_t,  \dots, \hat{h}^{n}_t}],\notag
\end{align}
where different levels have different convolution components $\text{Conv}_{l_i}$, while nodes at the same level share the same parameter $\Theta_{l_i}$.

It is important to emphasize that HTS is different from general graph-structured time series, where spatial-temporal information is passed between adjacent nodes, but the relationship between nodes in hierarchical structure only contains value aggregation, which indicates the message passing mechanism of graph time series (DCRNN~\cite{li2017diffusion} and STGCN~\cite{yu2017spatio}) is not appropriate for our problem. We compare it with our SLOTH method in Appendix E.

\subsubsection{\textbf{Bottom-Up Attention Module (BU-Attn)}}
\label{section: bottom_up}

This module integrates temporal features of nodes at the bottom levels to their ancestors at top levels to enhance the ability of adapting to dynamic pattern variations including sudden level or trend changes of time series.
% This module integrates temporal features of nodes at the bottom-levels to top-levels to enhance the variability of dynamic patterns which means more flexible and .

TD-Conv carries top-level information downward to improve the predictions at the bottom levels.  On the other hand, the information from the bottom-levels should also be useful for the prediction of the top-levels, due to their relationship of value aggregation.

Please note that feature aggregation is different from value aggregation (Eq. \eqref{eq_sum}). Specifically, the summing matrix ($\bS$) is actually a two-level hierarchical structure, rather than a tree structure that is reasonable for value aggregation, but causes structural information loss in feature aggregation. Direct summing operation is inappropriate for feature aggregation as there is no relationship of summation between parents and children in the feature space.

We therefore adopt the attention mechanism \cite{vaswani2017attention,nguyen2020treestructured} to aggregate temporal features from the bottom levels, due to the following considerations: 1) hierarchical structure has variations (different number of levels and children nodes); 2) child nodes contribute differently to parents with various scales/dynamic patterns. The attention mechanism is appropriate to aggregate the feature as it is flexible for various structures and can learn weighted contributions based on feature similarities.

\begin{align}
    \Tilde{h}^{\cup_l}_t& =\operatorname{Softmax}\left(\frac{\mathbf{Q}^{\cup_l}_t  (\mathbf{K}^{\cup_{l+1}}_t)^\mathsf{T}}{\sqrt{d_{h}}}\right) \mathbf{V}^{\cup_{l+1}}_t ,\label{eq:at-agg}\\
    \Tilde{\mathbf{H}}_t &= [\Tilde{h}^{\cup_{1}}_t,  \dots, \Tilde{h}^{\cup_{n_l}}_t ],\nonumber\\
    \mathbf{V}_t^{\cup_l} &= \Tilde{h}_t^{\cup_l} - \hat{h}^{\cup_{l}}_t + \bar{h}^{\cup_{l}}_t.\label{eq_value}
\end{align}

\begin{algorithm}[tb]
\caption{Bottom-up Attention Module}
\label{algo_attn}
\begin{algorithmic}[1]
    \REQUIRE $\mathbf{\bar{H}}_t, \mathbf{\hat{H}}_t, \phi_q, \phi_k $
    \STATE $l \gets n_l-1,  \Tilde{h}^{\cup_{n_l}}_t = \hat{h}^{\cup_{n_l}}_t, \mathbf{V}_t = 0 $
    \STATE $\mathbf{V}^{n_l}_t = \hat{h}^{\cup_{n_l}}_t $,  
    \STATE  \textit{/* $\cup_i $ means union of nodes at level i}
    \WHILE{$l > 0$}
        \STATE $\mathbf{Q}_t^{\cup_l} = F(\bar{h}^{\cup_l}_t; \phi_q)$
        \STATE $\mathbf{K}_t^{\cup_{l+1}} = F(\bar{h}^{\cup_{l+1}}_t; \phi_k)$
        \STATE Update $\Tilde{h}_t^{\cup_l}$ according to Eq.~\eqref{eq:at-agg}
        \STATE Update $\mathbf{V}_t^{\cup_l} = 
        \Tilde{h}_t^{\cup_l} - \hat{h}^{\cup_{l}}_t + \bar{h}^{\cup_{l}}_t $ 
        % \STATE Update $V_t^{\cup_l}$ based on Equation \ref{eq_value}
        \STATE $l \gets l-1,$
    \ENDWHILE
    \RETURN $\mathbf{\Tilde{H}}_t$
\end{algorithmic}

\end{algorithm}

% \begin{algorithm}[htbp]
% \caption{Bottom-up Attention Module}
% \label{algo_attn}
% \begin{algorithmic}
%     \REQUIRE $\mathbf{\bar{H}}_t, \mathbf{\hat{H}}_t, \phi_q, \phi_k $
%     \STATE $l \gets n_l-1,  \Tilde{h}^{\cup_{n_l}}_t = \hat{h}^{\cup_{n_l}}_t, \mathbf{V}_t = 0 $
%     \STATE $\mathbf{V}^{n_l}_t = \hat{h}^{\cup_{n_l}}_t $,  
%     \STATE  \textit{/* $\cup_i $ means union of nodes at level i}
%     \WHILE{$l > 0$}
%         \STATE $\mathbf{Q}_t^{\cup_l} = F(\bar{h}^{\cup_l}_t; \phi_q)$
%         \STATE $\mathbf{K}_t^{\cup_{l+1}} = F(\bar{h}^{\cup_{l+1}}_t; \phi_k)$
%         \STATE Update $\Tilde{h}_t^{\cup_l}$ according to Eq.~\eqref{eq:at-agg}
%         \STATE Update $\mathbf{V}_t^{\cup_l} = 
%         \Tilde{h}_t^{\cup_l} - \hat{h}^{\cup_{l}}_t + \bar{h}^{\cup_{l}}_t $ 
%         % \STATE Update $V_t^{\cup_l}$ based on Equation \ref{eq_value}
%         \STATE $l \gets l-1,$
%     \ENDWHILE
%     \RETURN $\mathbf{\Tilde{H}}_t$
% \end{algorithmic}
% \vspace{-1mm}
% \end{algorithm}
Specifically, as shown in Algorithm~\ref{algo_attn}, attention process starts from the second last to the first level (except for leaf nodes). Parent node takes the original temporal feature $\bar{h}_t$ to generate the query, and the corresponding child node takes the original feature $\bar{h}_t$ to generate the key. Child node feature value $\mathbf{V}_t$ is updated as attention process going upward as in Eq.~\eqref{eq_value}. 
$\Tilde{h}^{\cup_l}_t$ is the feature of all nodes at level $l$ by attention aggregation, which contains the contributions from both children and their ancestors. $\mathbf{V}_t^{\cup_l}$ is attention value of each node at level $l$, which is used in Eq.~\eqref{eq:at-agg}. Since we not only attempt to strengthen the information of children and the node's own features, but also to weaken the parents' influence in the bottom-up attention process because parent's feature becomes the node's own feature in the next iteration of BU-Attn, therefore, we subtract top-down temporal features $\hat{h}_t^{\cup_l}$ and add original temporal features $\bar{h}_t^{\cup_l}$.

It is important to note that computation process at same level can be executed concurrently because it is only related to self temporal hidden feature $\bar{\mathbf{H}}_t$. 
All the experiments show our bottom-up attention aggregation mechanism carries the bottom-level information containing dynamic patterns to top-level to improve the forecasting performance.
% An ablation study shows bottom-up attention mechanism will improve the coherency when without any explicit reconciliation after prediction. 
More importantly, both TD-Conv and BU-Attn can be independent components to be used in the fitting process (i.e., step $t$ of RNN) for better prediction accuracy, or to be used after the temporal patterns are obtained, trading accuracy for faster computation.

\subsubsection{\textbf{Base Forecast Module}}
This module serves as prediction generation based on dynamic features, and the prediction can either be probabilistic or point estimates. Our framework is agnostic to the forecasting models, such as MLP~\cite{gardner1998artificial}, seq2seq \cite{sutskever2014sequence}, and attention networks \cite{vaswani2017attention}.  
% In order to leverage td-conv and bu-attn mechanism completely, 
We employ the MLP to generate the base point estimates for its flexibility and simplicity.  
% We apply residual connection between temporal feature and forecasting module to avoid temporal feature loss after hierarchical learning as follows,
In order to avoid the loss of information of temporal features in the cascade of hierarchical learning, we apply residual connection between temporal feature extraction module and base forecast module as follows
% And same as top-down convolution, our approach also add gate mechanism to  avoid temporal feature lost.
\begin{equation}
\label{eq:attn_filter}
    {z}_t = \sigma(MLP(\Tilde{h}_t));
    h_t = (1 - {z}_t) \Tilde{h}_t + {z}_t \bar{h}_t
\textrm{ . }
\end{equation}
Then we apply MLP to generate base forecasts as follows 
\begin{equation}
    \label{eq:pred}
    \hat{y}^i_t = MLP(h^i_t), \quad \hat{\by}_t = [\hat{y}^1_t,  \dots, \hat{y}^{n}_t].\footnote{Note that all nodes share the same generation model.}
\end{equation}

\subsection{Task-based Constrained Optimization}
% As mentioned before, hierarchical time series forecasting task requires the prediction result satisfy aggregation summing constraint as Eq ~\ref{eq_sum} shows. 
In this section, we introduce a task-based optimization module that leverages the deep neural optimization layer to achieve targets in realistic scenarios, while satisfying coherence and task-based constraints.  

\subsubsection{\textbf{Optimization with Coherence Constraint in Forecasting Task}}
We formally define HTS forecasting task as a prediction and optimization problem in this section.  As shown in Eq.~\eqref{eq:recon}, 
% \textcolor{red}{ can't tell why 16 instead of 12 below }
reconciliation on base forecasts can be represented as a constrained optimization problem ~\cite{rangapuram2021end}, where two categories of constraints are considered, i.e., the equality constraints representing coherency, and the inequality constraints 
ensuring the reconciliation is restricted, which means the adjustment of the base forecasts is limited in a specific range to reduce the deterioration of forecast performance,
\begin{equation}
    \label{eq:recon}
    \begin{aligned}
      &\Tilde{\by}_t = \mathop{\arg\min}_{\by \in \mathbb{R}^n} \|\by - \hat{\by}_t \|_2
      = \mathop{\arg\min}_{\by \in \mathbb{R}^n} \frac{1}{2}\by^{\mathsf{T}}\by - \hat{\by}_t^{\mathsf{T}}\by \\
      &\text{subject to} 
      \begin{cases}
      \mathbf{A}{\by} = \mathbf{0},\\
    %   {\by} \ge \mathbf{0}, \\
       \delta_1 abs(\hat{\by}_t) - \varepsilon_1 \le \by - \hat{\by}_t \le \delta_2 abs(\hat{\by}_t) + \varepsilon_2,
      \end{cases}
     \end{aligned}
\end{equation}
where $\hat{\by}$ is the base forecasts without reconciliation, and $\delta_i,\varepsilon_i, i=1,2$ are some predefined constants.

Recall that the aforementioned end-to-end optimization architecture~\cite{rangapuram2021end} provides a closed-form solution for reconciliation problem. It projects the base forecasts into the solution space effectively by multiplying reconciliation matrix, which only depends on aggregation matrix $\mathbf{S}$ and is thus easy to calculate (for convenience, the details are supplied in Appendix A). 
However, this procedure only considers aggregation constraints without limiting adjustment scale which may sometime cause the reconciled results $\Tilde{\by}$ become unreasonable, e.g., a negative value for small base forecasts in demand forecasting. 
In addition, loss based on reconciliation projection derails gradient magnitudes and directions, which may cause the model not to converge to the optimum.  
Moreover, it is not a general solution for real-world scenarios, where more complex task-related constraints and targets are involved.

% To include non-negative constraints and keep training efficient, 
To keep reconciliation result reasonable and training efficient,
we utilize neural network layer OptNet to solve the constrained reconciliation optimization problem, which is essentially a quadratic programming problem. The Lagrangian of formal quadratic programming problem is defined in Eq.~\eqref{eq: optnet} \cite{amos2017optnet}, where equality constraints are $\mathbf{A}\mathbf{z} = \mathbf{b}$ and inequality constraints are $\mathbf{G}\mathbf{z} \le \mathbf{h}$:
% \begin{small}
\begin{equation}
    \label{eq: optnet}
    L(\mathbf{z}, {\nu}, {\lambda}) = \frac{1}{2}\mathbf{z}^{\mathsf{T}}\mathbf{Q}\mathbf{z} - \mathbf{q}^{\mathsf{T}}\mathbf{z} + \nu^{\mathsf{T}}(\mathbf{A}\mathbf{z}-\mathbf{b}) + \lambda^{\mathsf{T}}(\mathbf{G}\mathbf{z}-\mathbf{h}).
\end{equation}
% \end{small}
When applied to hierarchical reconciliation problems (where we take the special range constraint $\by \ge \mathbf{0}$ as an example), 
the Lagrangian can be revised to 
\begin{equation}
    \label{eq: hier_optnet}
    L(\by, \nu, \lambda) = \frac{1}{2}\by^{\mathsf{T}}\by - \hat{\by}^{\mathsf{T}}\by + \nu^{\mathsf{T}}\mathbf{A}\by + \lambda^{\mathsf{T}}(-\mathbf{I}\by),
\end{equation}
where $\nu,\lambda$ are the dual variables of equality and inequality constraints respectively. Then we can derive the differentials of these variables according to the KKT condition, and apply linear differential theory to calculate the Jacobians for backpropagation. The detail is as follows

\begin{equation}
    \resizebox{1.\linewidth}{!}{$
    \begin{bmatrix}
    \mathbf{I} & -\mathbf{I} & 
    \mathbf{A}^\mathsf{T} \\
    diag(\lambda)(-\mathbf{I}) & diag(-\by) & \mathbf{0} \\
    \mathbf{A} & \mathbf{0} & \mathbf{0}
    \end{bmatrix}
    \begin{bmatrix}
    d\by  \\
    d\lambda \\
    d\nu
    \end{bmatrix}
    = -
     \begin{bmatrix}
    d\by - d\hat{\by} - d\lambda + d\mathbf{A}^{\mathsf{T}}\nu \\
    - diag(\lambda)d\by \\
    d\mathbf{A} \by 
    \end{bmatrix}
    \mathrm{,}$}
\end{equation}
where function $diag(\cdot)$ means diagonal matrix. We can infer the conditions of the constrained reconciliation optimization problem from the left side, and compute the derivative of the relevant function with respect to model parameters from the right side. In practice, we apply OptNet layer to obtain the solution of argmin differential QPs quickly to solve the linear equation. In this way, our framework achieves end-to-end learning by directly generating reconciliation optimization results, while calculating the derivative and backpropagating the gradient to the optimization model automatically. 

\subsubsection{\textbf{Optimization with Task-based Constraints and Target for Real-World Scenarios}}
The coherence constraint is enough in forecasting tasks when the only concern is prediction accuracy, which is, however, most likely unrealistic in real-world tasks with specific limitations and practical targets. Such tasks can be further revised as follows
\begin{equation}
    \label{eq:recon_task}
    \begin{aligned}
      &\mathcal{J}(\hat{\by}) = \mathop{\arg\min}_{\by}{f(\hat{\by}, \by)}\\
    \text{subject to} 
    &\begin{cases}
    \mathbf{A}\by = \mathbf{0}, e_j = 0, \;j=1, \dots, n_{\text{eq}},\\
    g_i(\by, \hat{\by}) \le 0, i=1, \dots, n_{\text{ineq}},    
    \end{cases}
     \end{aligned}
\end{equation}
where $f$ is the task-based quadratic objective, $e_j$ represents task-based equality constraint other than coherence constraint, $n_{\text{eq}}$ is the number of equality constraints, and $g_i$ is an inequality constraint, $n_{\text{ineq}}$ is the number of task-based inequality constraints. 
% And hierarchical tasks in real-world always have more realistic constraints and optimization targets,
Eq.~\eqref{eq:recon_task} can be efficiently solved using differential QP layer in an end-to-end fashion \cite{donti2017task},  where we need to transform our target into a quadratic loss and add equality/inequality constraints.  We construct a scheduling experiment on M5 dataset in the following section to validate the superiority of our framework on realistic tasks.

\section{Experiments}
\begin{table}[t]
    \centering
    \resizebox{0.8\linewidth}{!}{
    \begin{tabular}{|c|c|c|c|c|c|}
        \hline
         Dataset & levels & nodes & structure & freq \\ \hline
         Labour & 4 & 57 & 1, 8, 16, 32 &  1M \\ \hline
         Tourism & 4 & 89 & 1, 4, 28, 56 & 3M \\ \hline
         M5 & 5 & 114 & 1, 3, 10, 30 ,70 & 1D\\ \hline
    \end{tabular}}
    \caption{Dataset statistics. Structure column shows the number of nodes at each level from top to bottom, as for frequency (`freq'), 1D means one day and 3M means three months.}
    \label{tab:dataset_statisitic}
\end{table}
\newcolumntype{s}{>{\hsize=1.4\hsize\small}X}
\begin{table*}[htbp]
    \centering
    \resizebox{.9\linewidth}{!}{
    \begin{tabularx}{\textwidth}{s|XX|XX|XX}
        \hline
         \textbf{Model}  & \multicolumn{2}{X}{Tourism} & \multicolumn{2}{X}{Labour}  & \multicolumn{2}{X}{M5} \\
         \hline
         Metric & MAPE & w-MAPE & MAPE & w-MAPE & MAPE & w-MAPE \\
         \hline
         ARIMA-BU & 0.2966 & 0.1212 &0.0467 & 0.0457 & 0.1134 & \textbf{0.0638}\\
         ARIMA-MINT-SHR & 0.2942 & 0.1237 & 0.0506 & 0.0471 & 0.1140 & 0.0675  \\
         ARIMA-MINT-OLS  & 0.3030 & 0.1254  & 0.0505 & 0.0467 & 0.1400 & 0.0733 \\
         ARIMA-ERM & 1.6447 & 0.6198 & 0.0495 & 0.0402 & 0.1810 & 0.1163 \\
         PERMBU-MINT& 0.2947(0.0031) & 0.1057(0.0004)  & 0.0497(0.0003) & 0.0453(0.0002) & 0.1176(0.0005) & 0.0759(0.0007)\\
         \hline
         DeepAR-Proj & 0.3214(0.0202) & 0.1171(0.0116) & 0.0423(0.0016) & 0.0290(0.0013)  
         &0.1546(0.0165) & 0.0951(0.0195) \\
         DeepVAR-Proj & 0.4214(0.0548) & 0.2162(0.0307)  & 0.0936(0.0206) & 0.0884(0.0235) &
         0.2019(0.0279) & 0.1615(0.0311) \\
         NBEATS-Proj & 0.3295(0.0231) & 0.1359(0.0264)  & 0.0355(0.0018) & 0.0268(0.0043) 
         &0.2256(0.0399) & 0.1952(0.0637) \\
         INFORMER-Proj & 0.5401(0.0339) & 0.5566(0.0261)  & 0.1537(0.0685) & 0.1455(0.0683) 
         &0.3123(0.0735) & 0.3098(0.0708) \\
         AUTOFORMER-Proj & 0.3983(0.0678) & 0.1862(0.0596)  & 0.0455(0.0037) & 0.0367(0.0016) 
         &0.1654(0.0153) & 0.1308(0.0560) \\
         FEDFORMER-Proj & 0.3741(0.0291) & 0.1685(0.0180)  & 0.0440(0.0038) & 0.0334(0.0024) 
         &0.1505(0.0139) & 0.1188(0.0044) \\
         DeepAR-BU & 0.3065(0.0123) & 0.1154(0.0097)  & 0.0378(0.0014) & 0.0278(0.0022)
         & 0.1151(0.0017) & 0.0686(0.0012) \\
         DeepVAR-BU & 0.4135(0.0562) & 0.2195(0.0370)  & 0.1112(0.0371) & 0.1008(0.0352) 
         & 0.1851(0.0153) & 0.1494(0.0140) \\
         NBEATS-BU & 0.2904(0.0308) & 0.1259(0.0183)  & 0.0393(0.0031) & 0.0310(0.0046)
         &0.1740(0.0221)  & 0.1398(0.0294) \\
         INFROMER-BU & 0.5694(0.0065) & 0.5707(0.0072)  &0.1654(0.0824) & 0.1580(0.0840) & 0.3128(0.0728)  &0.3099(0.0706) \\
         AUTOFORMER-BU & 0.3787(0.0578) & 0.1868(0.0084)  & 0.0519(0.0034) & 0.0505(0.0011) 
         &0.1506(0.0146) & 0.1143(0.0049) \\
         FEDFORMER-BU & 0.3408(0.0099) & 0.1544(0.0097)  & 0.0464(0.0041) & 0.0369(0.0027) 
         &0.1424(0.0141) & 0.1081(0.0034) \\
         \hline
         \textbf{SLOTH(Opt)(ours)} & 0.2613(0.0017) & 0.1032(0.0012) & \textbf{0.0328(0.0006)} & \textbf{0.0183(0.0008)} & \textbf{0.1116(0.0018)} & 0.0696(0.0017) \\
          SLOTH(Proj) & 0.2780(0.0051) & 0.1098(0.0008)  & 0.0370(0.0052) & 0.0228(0.0072)
          & 0.1121(0.0014) & 0.0704(0.0005) \\
          SLOTH(BU) & \textbf{0.2583(0.0015)} & \textbf{0.0991(0.0021)}  & 0.0391(0.0051) & 0.0248(0.0065) 
          & 0.1127(0.0017)& 0.0703(0.0023) \\
         \hline
    \end{tabularx}}
    \caption{MAPE and weighted-MAPE (w-MAPE) metric values over five independent runs for baselines such as traditional reconciliation methods and end-to-end methods, as well as our approach. The value in brackets is the variance over the five runs. }
    \label{tab:main_results}
\end{table*}

In this section, we conduct extensive evaluations on real-world hierarchical datasets. Firstly, we evaluate the performance of our framework, and compare our approach against the traditional statistical method and end-to-end model (HierE2E). We then add more practical constraints to M5 dataset, building a meaningful optimization target to solve an inventory management problem, and again evaluate various approaches for hierarchical tasks under these realistic scenarios. 
% \replace{Finally, we apply our framework to a real-world task in cloud resource scheduling of a giant internet company shown in Appendix E}{We also apply our framework to a real-world task in cloud resource scheduling of a leading internet company shown in Appendix E}.
Our framework on a real-world task of cloud resource scheduling in Ant Group is shown in Appendix E.

\subsection{Real-world datasets}
We take three publicly available datasets with standard hierarchical structures.
\begin{itemize}
\item{
Tourism \cite{bushell2001tourism,athanasopoulos2009hierarchical} includes an 89-series geographical hierarchy with quarterly observations of Australian tourism flows from 1998 to 2006, which is divided into 4 levels.  Bottom-level contains 56 series, aggregated-levels contain 33 series, prediction length is 8.
This dataset is frequently referenced in hierarchical forecasting studies
\cite{hyndman2018forecasting}. }
% \cite{hyndman2018forecasting,taieb2021hierarchical}. }
\item{
Labour \cite{australiaLabour} includes monthly Australian employment data from Feb. 1978 to Dec. 2020. By using included category labels, we construct a 57-series hierarchy, which is divided into 4 levels. Specifically, bottom-level contains 57 series, aggregated-levels contain 49 series in total, and prediction length is 8.}

\item M5 \cite{han2021simultaneously} dataset describes daily sales from Jan. 2011 to June 2016 of various products. We construct 5-level hierarchical structure as state, store, category, department, and detail product from the origin dataset, resulting in 70 bottom time series and 44 aggregated-level time series. The prediction length is 8.
\end{itemize}

\subsection{Results Analysis}
\label{sec:m5_sched}
\begin{figure*}
    \centering
    \includegraphics[width=.9\linewidth]{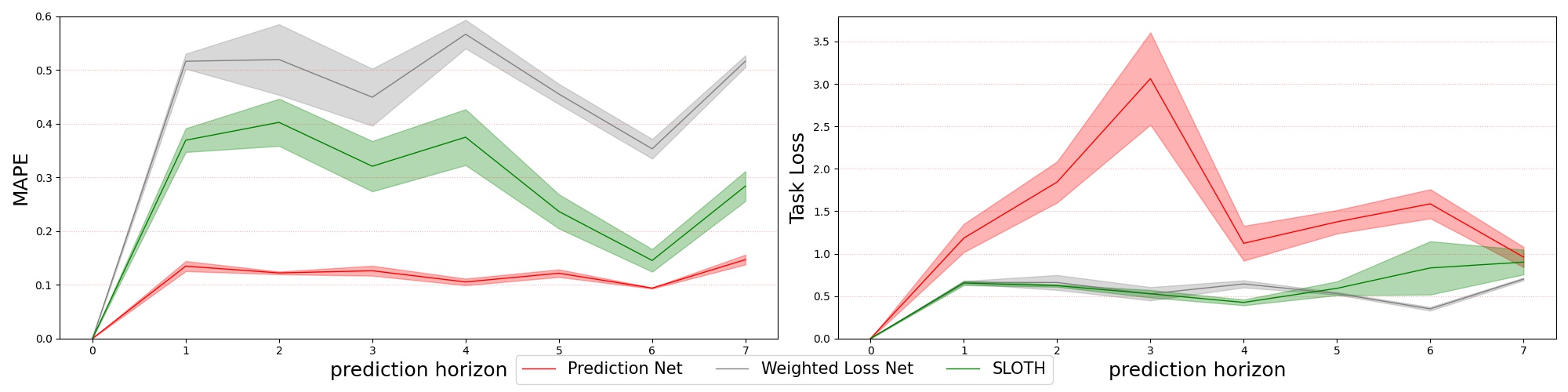}
    \caption{Results of 5 independent runs of M5 scheduling problem for 7-day prediction (lower MAPE and lower task loss are better). As expected, the network using prediction loss achieves the highest accuracy.  Our task-net improves the prediction performance by 36.8\% compared to weighed loss net, and outperforms the prediction loss net by 53.2\%.
    }
    \label{fig:inv_uni}
\end{figure*}
In this section, we validate the overall performance of our method in the prediction task on three public datasets. We report scale-free metrics \textit{MAPE} and scaled metrics \textit{weighted-MAPE} to measure the accuracy.  We apply several representative state-of-the-art forecasting baselines (details shown in Appendix B), including DeepAR \cite{salinas2020deepar}, DeepVAR \cite{salinas2019high}, NBEATS \cite{oreshkin2019n}, and former-based methods \cite{zhou2021informer,wu2021autoformer, zhou2022fedformer}. We then combine these methods with traditional bottom-up aggregation mechanism and closed-form solution \cite{rangapuram2021end} (DeepVar-Proj is HierE2E) to generate reconciliation results.

Table~\ref{tab:main_results} reports the results. The top part shows results of traditional statistic methods, the middle part is of deep neural networks methods with closed-formed solution and bottom-up reconciliation, and the bottom part is the results from our approach and the combination of our forecasting mechanism with traditional bottom-up (BU) and closed-form projection methods (Proj).
% \textcolor{red}{?? fore}.

We can see that traditional statistic methods perform poorly compared with deep neural networks. 
NBEATS performs best on Tourism dataset. In particular, NBEATS-BU performs best on MAPE while NBEATS-Proj performs best on weighted-MAPE. However, informer-related methods perform poorly. One possible explanation is that it requires much larger training dataset.
 
One can observe that the models performing best on MAPE do not perform as well on weighted-MAPE, which is caused by different level contributions on the overall performance, e.g., the bottom-level contributes more on MAPE but higher levels contribute more on weighted-MAPE.

% \textcolor{red}{I am not sure whether the modification is correct}

Our proposed approach (SLOTH) has superior performance compared to other methods on both MAPE and weighted-MAPE metrics in most scenarios. Specifically, SLOTH achieves best performance among all models on Labour and M5 datasets, and ranks the second on Tourism dataset and the third on w-MAPE for M5.  Besides optimization reconciliation, we also combine our forecasting mechanism with the aforementioned bottom-up and closed-form projection methods, and both methods achieve higher accuracy than the baselines.  Even SLOTH-BU, with primitive structure, achieves the best performance in Tourism dataset. Please note that smaller variances indicate that our framework shows stabler performances across various scenarios. 

In conclusion, our SLOTH mechanism improves the forecasting performance, and optimized reconciliation generates reasonable coherent predictions without much performance loss. We also assess the gains in the performance across all levels and running time in Appendix D, and the ablation study for each component is presented in Appendix E.

\subsection{M5 Scheduling Task}
In this section, we apply our framework to realistic scenarios with meaningful task-based constraints and targets, using a designed scheduling task for product sales based on M5 dataset.  
Specifically, we define a meaningful task that minimize the cost of scheduling and inventory with more practical conditions: 
\begin{itemize}
    \item Underestimation and overestimation contribute differently to the final cost. Underestimation implies that the store needs to order more commodity to fulfill the demand, which increases scheduling cost. Overestimation implies that the store has to keep the extra commodity, which increases inventory cost.
    \item Different levels generate different weighted contribution due to aggregation, e.g., scheduling and inventory costs for the top-levels are less than those for the bottom-levels 
    % \replace{because the demand is higher at the top-levels.}
    {because the company scale is larger at the top-levels.}
    \item Commodities of different types have different inventory and scheduling costs, since the shelf life for food products is shorter than household products, i.e., the inventory cost is lower for shorter storage time and the scheduling cost is higher due to the need for faster transportation.
\end{itemize}

\textbf{Settings.} We assume scheduling takes place every week.  We set prediction length to 7 and context length to 14. The other penalty settings and target are detailed in Appendix F.  We then compare the outcome from the following models:
\begin{enumerate}
    \item Prediction Net: prediction model that takes the prediction metric (MAE) as the loss for optimization, and bottom-up approach for coherency.
    \item Weighted Loss Net: prediction model that takes task cost (weighted-MAE) as the loss for optimization, and bottom-up for coherency.
    \item SLOTH: our end-to-end approach that takes both cost and constraints, with the task loss for optimization.
\end{enumerate} 

As shown in Figure~\ref{fig:inv_uni}, the prediction Net model performs the best on prediction (MAPE), which is the training objective.  As for the task cost, which the scheduler really cares in practice, our SLOTH framework outperforms the Prediction Net by a large margin. Specifically, our model improves the task-cost performance by 36.8\% compared to the Prediction Net, while at the same time achieving a similar task loss target as the Weighted Loss Net, but with an improvement of 53.2\% in the prediction accuracy.

\section{Conclusion}
In this paper, we introduced a novel task-based structure-learning framework (SLOTH) for HTS. We proposed two tree-based mechanisms to utilize the hierarchical structure for HTS forecasting. The top-down convolution integrates the temporal feature of the top-level to enhance stability of dynamic patterns, while the bottom-up attention incorporates the features of the bottom-level to improve the coherency of the temporal features.  In the reconciliation step, we applied the deep neural optimization layer to produce the controllable coherent result, which also accommodates complicated realistic task-based constraints and targets under coherency without requiring any explicit post-processing step.  We unified the goals of forecasting and decision-making and achieved an end-to-end framework.  We conducted extensive empirical evaluations on real-world datasets, where the competitiveness of our method under various conditions against other state-of-the-art methods were demonstrated. 
Furthermore, our ablation studies proved the efficacy of each component we designed.
Our method has also been deployed in the production environment in Ant Group for its cloud resources scheduling.

\bibliography{aaai23}

\clearpage
\appendix

\section{Appendix A Related Work}

\label{sec:related_work}
\subsection{Hirarchical Time Series Prediction}

\subsubsection*{\textbf{Traditional approaches}} 
Traditionally, coherence is utilized as a tool helping generate base forecasts. Base forecasts of a pre-specified level in the hierarchy are generated first, then according to the coherence, reconciliation methods are applied to generate base forecasts of other levels. There are three types of approaches following this routine: 
\textbf{Bottom-Up mechanism} \cite{gross1990disaggregation}
generates forecasts at the bottom level and then aggregates them according to the hierarchical structure to produce high level forecasts as in Eq.~\eqref{eq_sum}, where
hierarchical structure information was utilized explicitly during the aggregations.  The major shortcoming of this approach is that the bottom-level series tend to be noisy and there is a high risk of inaccurate forecasting~\cite{athanasopoulos2017forecasting,wickramasuriya2019optimal}, which accumulates during aggregation.
To overcome this weakness,  a \textbf{Top-Down mechanism} was proposed in \cite{athanasopoulos2009hierarchical}, where the top-level time series are first forecasted and then disaggregated to bottom-levels. Forecasting at the top-level is an easier task because the temporal pattern is normally more stable and less noisy, but the disaggregation procedure still cannot fully utilize the information of the hierarchical structure.  Considering the disadvantages of aforementioned approaches, a compromised approach called \textbf{Middle-out mechanism} is proposed in ~\cite{athanasopoulos2017forecasting}, where forecasts are first produced at a middle level of the hierarchy, followed by aggregations for the higher levels and disaggregations for the lower ones.  A significant weakness of these methods lies in information loss, i.e., the characteristics of time series at other levels are not integrated, which worsens with deeper hierarchies \cite{Kourentzes2019Cross}.

\subsubsection*{\textbf{Forecast Reconciliation}}
In order to overcome the weaknesses of the above approaches, a two-stage procedure is proposed to produce coherent forecasts for all series in the hierarchy: 
\begin{enumerate}
    \item Produce forecasting results for time series of each level independently without considering the structure.
    \item Conduct reconciliation on base forecasts using the hierarchy structure, to obtain the coherent forecasts.  
\end{enumerate}

Given the $h$-step-ahead base forecasts $\hat{\by}_{T+h}$, the reconciliation step \cite{Hyndman2011Optimal,wickramasuriya2019optimal} obtains coherent forecasts as follows:
\begin{equation}
\label{eq_spy}
\tilde{\by}_{T+h}=\bS\bP\hat{\by}_{T+h},
\end{equation}
where matrix $\bP \in \mathbb{R}^{m \times n}$ projects base forecasts of all nodes to those at the bottom level, $\bS \in \{0,1\}^{n \times m}$ is the aggregation matrix, and $\tilde{\by}_{T+h}$ is the coherent forecasts.  Aggregation matrix ensures that reconciliation forecasts $\tilde{\by}_{T+h}$ satisfy the coherence constraint. The key to the reconciliation step is to solve for the matrix $\bP$, and the Trace Minimization (MinT) algorithm~\cite{wickramasuriya2019optimal} provides the optimal solution of the analytical expression 
$\bP=(\bS^\mathsf{T}\mathbf{W}^{-1}_h\bS)^{-1}$ $(\bS^\mathsf{T}\mathbf{W}^{-1}_h)$,
where $\mathbf{W}_{h}$ is the covariance matrix of the $h$-step-ahead forecast error $\hat{\varepsilon}_{T+h}=\by_{T+h}-\hat{\by}_{T+h}$, and the empirical risk minimization (ERM) algorithm~\cite{ben2019regularized} puts unbiased error and variance error into the objective function and then derive the best reconciliation result through solving an ERM problem.

There are several major issues with these two-stage procedures: 
1) These methods employ univariate time series models as the base forecaster predominantly, such as linear auto-regressive (AR) models, which are initially trained independently without considering the relationships across time series; 
2) The base forecasts are revised without any regard to the learned model parameters; 3) Several methods assume that the individual base forecast is unbiased, which is inconsistent with many real-world tasks.

An end-to-end framework for coherent probabilistic forecasting is proposed recently in \textbf{Hier-E2E} \cite{rangapuram2021end},
leveraging a closed-form projection from the space of base forecasts to the space of coherent solution with minimization of revision
% \comment{minimization of revision}{I don't quite get the point here...} 
,in order to retain the prediction as much as possible while satisfying the coherence constraint as
\begin{equation}
    \label{eq:proj}
    \tilde{\by}_{t+h} = \mathbf{M} \hat{\by}_{t+h} = [\mathbf{I} - \mathbf{A}^\mathsf{T} (\mathbf{A}\mathbf{A}^\mathsf{T})^{-1}\mathbf{A}] \hat{\by}_{t+h},
\end{equation}
where $\mathbf{A}$ is the constraint matrix as shown in Eq. \eqref{eq:coherent_co}.  The closed-form projection matrix $\mathbf{M}$ can be calculated before training since it only depends on hierarchical matrix $\mathbf{A}$.  This procedure improves the reconciliation efficiency without incurring any post-processing penalty.  Another advantage of this method is the use of multivariate model DeepVAR \cite{salinas2019high} to produce base forecasts, where training takes all time series simultaneously, followed by DNN to extract relationship between nodes, which leads to better forecast accuracy.  However, this method has the following limitations: 1) the structural information in the hierarchy is not explicitly incorporated to produce the base forecasts; 2) the closed-form projection ignores the inequality constraints;
 and 3) more general and complex task-related constraints beyond the coherence constraint are not considered.

To the best of our knowledge, none of the existing approaches takes hierarchical structure information explicitly into consideration in base forecasting.  Inspired by structure learning \cite{nguyen2020treestructured, mou2014tbcnn, li2017diffusion, yu2017spatio}
%\textcolor{red}{Neet reference}
\add{,} 
we propose dual tree-based mechanisms to incorporate hierarchical structure into temporal features for more accurate base forecasts.  Furthermore, all the aforementioned methods aim at ensuring the coherence, ignoring task-based targets or realistic constraints, making them impractical for real-world applications.  Our optimization reconciliation approach, on the other hand, takes the final target as an optimization goal while satisfying task-related constraints including coherence constraint.

\subsection{Prediction and Optimization}
The intersection of prediction and decision models has been a prominent topic in recent years \cite{kotary2021end}, 
where decision models can be partially represented as constrained optimization problems. 
% Different from traditional methods where the optimization process is done when the prediction is finished where the optimization will not influence optimization, end-to-end solutions have been proposed to solve such tasks where the deep neural network as a layer has been employed to solve the optimization problem, 
\begin{equation}
\begin{aligned}
    \mathcal{J}(\mathbf{y}) =& \mathop{\arg\min}_{\mathbf{z}} f(\by, \mathbf{z}) \\
    \text{subject to} &\; \mathbf{z} \in C_\mathbf{z},
\end{aligned}
\end{equation}
where $\by$ is the prediction results generated by the time series model, $f({\by}, \cdot)$ is the objective function and $C_{\mathbf{z}}$ is the feasible solution space under specific constraints.  The training process is to find the optimal solution $\hat{\by}$ by minimizing the target loss $J(\by)$.
% \textcolor{red}{The following part is almost exactly the same as in the main text.  Better use a reference to the section}
The optimization process can be seen as a Quadratic Programming (QP) problem if target loss function is quadratic. 
Amos and Kolter \cite{amos2017optnet} employed GPU-ready QP solver (OPTNET) to solve constrained QP problems with a deep neural layer.  In the training process, OPTNET derives argmin gradient for backpropagation by differentiating Lagrange's KKT conditions using matrix differential calculus, and generates the gradient to optimize implicit inference layer.  Optimization result is obtained in forward pass. 
Donti et al.~\cite{donti2017task} then proposed a predict-and-optimize architecture by applying the QP solver to time series tasks with task-specified constraints.
% These previous works provide a solution to hierarchical time series forecasting tasks in realistic scenarios, which will generate the best task-based results while satisfying coherency constraint and other realistic constraints.
% Both works provide an end-to-end learning  scheme to HTS forecasting tasks in realistic scenarios, generating the best task-based results while satisfying coherency and other realistic constraints.

\section{Appendix B Baselines}
\label{sec:baselines}
The baselines for prediction and reconciliation is as follows, we include traditional and deep learning methods.
\begin{itemize}
    \item Traditional statistical models \footnote{All methods can be found in hts R package: \url{https://cran.r-project.org/web/packages/hts/vignettes/hts.pdf}}
    \begin{itemize}
       \item ARIMA-BU: ARIMA is a standard statistic prediction method, and bottom-up reconciliation (BU) is the basic reconciliation method.
        \item ARIMA-MINT-SHR: ARIMA prediction with min-trace reconciliation using \textit{SHR} mechanism to generate covariance \cite{wickramasuriya2019optimal}.
        \item ARIMA-MINT-OLS:ARIMA prediction with min-trace reconciliation method using \textit{OLS} mechanism to generate covariance \cite{wickramasuriya2019optimal}.
        \item ARIMA-ERM: ARIMA prediction, using \textit{ERM} \cite{ben2019regularized} to generate reconciliation result.
        \item PERMBU-MINT: probabilistic hierarchical forecasting method with min-trace reconciliation \cite{taieb2017coherent}
    \end{itemize}
    \item Deep learning baselines are the combination of four baseline deep forecasting models with two reconciliation methods.
    \begin{itemize}
        \item
            \begin{itemize}
                \item DeepAR: an approach to produce accurate probabilistic forecasts, based on autoregressive recurrent neural network model, which is widely applied in industrial forecasting tasks. \footnote{https://github.com/arrigonialberto86/deepar} \cite{salinas2020deepar}
                \item DeepVar: 
                an representative RNN-based multi-variate probabilistic time series model with a Gaussian copula process to handle multi-variate Gaussion joint distributions. \footnote{https://github.com/zalandoresearch/pytorch-ts/tree/master/pts/model/deepvar} \cite{salinas2019high}
                \item NBEATS: 
                a deep neural architecture based on backward and
                forward residual links and a deep stack of neural layers to achieve interpretability. \footnote{https://github.com/philipperemy/n-beats} \cite{oreshkin2019n}
                \item INFORMER: an efficient transformer-based model in long sequence time-series forecasting with ProbSparse self-attention, self-attention distilling and generative style decoder.  \footnote{https://github.com/zhouhaoyi/Informer2020} \cite{zhou2021informer}
            \end{itemize}
        \item Reconciliation methods
        \begin{itemize}
            \item Bottom-Up (BU): a traditionally popular reconciliation method, used by most of the state-of-the-art methods as discussed in Appendix A. 
            \item Closed-form projection operator (Proj): the state-of-the-art end-to-end reconciliation method, with the closed-form projection minimizing the revision of base forecast while ensuring coherence. It is combined with DeepVar forecasting to form the best model in hierarchical forecasting in \textbf{HierE2E} \cite{rangapuram2021end}
        \end{itemize}
    \end{itemize}
\end{itemize}

\section{Appendix C Evalution Metrics}
We use \textit{Mean Absolute Percent Error} (MAPE) as the prediction metric, and adopt the weighted mape ($w\_MAPE$) to evaluate the performance when taking scaled differences across all levels into consideration:
\begin{equation}
\label{eq:mape}
    MAPE = \frac{100\%}{n} \sum_{i=1}^n \left|\frac{y^i - \hat{y}^i}{\hat{y}^i}\right|,
\end{equation}
\begin{equation}
\label{eq:weighted_mape}
    w\_MAPE = \sum_{i=1}^n \frac{\hat{y}^i\times|y^i - \hat{y}^i|}{\sum_{j=1}^n \hat{y}^j}.
\end{equation}
In the ablation study, we use $CO\_MAPE$ to measure the coherence satisfaction as shown in Eq.~\eqref{eq:co_mape}, and we do not consider the leaf node when calculating $CO\_MAPE$ because it does not contain any aggregation:
\begin{equation}
\label{eq:co_mape}
    CO\_MAPE = \frac{1}{r} \sum_{i=1}^r \frac{|y^i - \sum_{j \in c(i)}y^j|}{y^i},
\end{equation}
where $c(i)$ means the children of node $i$, and $r$ is number of all nodes minus the number of leaf nodes. 

\section{Appendix D Detailed Experiments Results }
\newcolumntype{m}{>{\hsize=5\hsize\scriptsize}X}
\begin{table*}
    \scriptsize
    \begin{tabularx}{\textwidth}{mm|cccc|cccc|ccccc}
        \hline
         \multicolumn{2}{c|}{Dataset} & \multicolumn{4}{c|}{Tourism} & \multicolumn{4}{c|}{Labour} & \multicolumn{5}{c}{M5} \\ \hline
        \multicolumn{2}{l|}{} & \multicolumn{4}{c|}{Level} & \multicolumn{4}{c|}{Level} &
        \multicolumn{5}{c}{Level}\\ \hline
        Model & Metric & 1 &2 &3 &4 & 1 &2 &3 &4 & 1 &2 &3 &4 & 5 \\ \hline
         \multirow{2}{0.08\textwidth}{ARIMA-BU} & MAPE & 
         0.0632 & 0.1055 & 0.2442 & 0.3406 &
         0.0439 & 0.0375 & 0.0391 & 0.0529 &
         0.0259 & 0.0498 & 0.0576 & 0.0851 & 0.1374 \\
         & w-MAPE & 
         0.0158 & 0.0248 & 0.0360 & 0.0444 &
         0.0109 & 0.0111 & 0.0112 & 0.0123 &
         0.0051 & 0.0098 & 0.0119 & 0.0159 & 0.0208\\ \hline
         \multirow{2}{0.08\textwidth}{PERMBU-MINT} & MAPE & 
         0.0571 & 0.1014 & 0.2450 & 0.3390 &
         0.0448 & 0.0422 & 0.0425 & 0.0556 &
         0.0423 & 0.0672 & 0.0705 & 0.0900 & 0.1393 \\
         & w-MAPE & 
         0.0142 & 0.0205 & 0.0324 & 0.0405 &
         0.0112 & 0.0108 & 0.0109 & 0.0123 &
         0.0086 & 0.0135 & 0.0148 & 0.0183 & 0.0225\\ \hline
         \multirow{2}{0.08\textwidth}{DEEPAR-BU} & MAPE & 
         0.0686 & 0.1034 & 0.2661 & 0.3455 &
         0.0236 & 0.0253 & 0.0265 & 0.047 &
         0.0353 & 0.0569 & 0.0649 & 0.0904 & 0.1365 \\
         & w-MAPE & 
         0.0172 & 0.0212 & 0.0337 & 0.0395 &
         0.0059 & 0.0065 & 0.0068 & 0.0086 &
         0.007 & 0.0112 & 0.0133 & 0.0167 & 0.0204\\ \hline
         \multirow{2}{0.08\textwidth}{DEEPAR-Proj} & MAPE & 
         0.0771 & 0.1161 & 0.2825 & 0.36 &
         0.024 & 0.0282 & 0.03 & 0.0526 &
         0.0615 & 0.0785 & 0.0917 & 0.1176 & 0.1842 \\
         & w-MAPE & 
         0.0193 & 0.0225 & 0.0349 & 0.0405 &
         0.006 & 0.0068 & 0.0072 & 0.009 &
         0.0123 & 0.0159 & 0.0186 & 0.0225 & 0.0258\\ \hline
         \multirow{2}{0.08\textwidth}{DEEPVAR-BU} & MAPE & 
         0.1672 & 0.1904 & 0.363 & 0.459 &
         0.0984 & 0.102 & 0.1035 & 0.1196 &
         0.1254 & 0.1333 & 0.1436 & 0.169 & 0.2011 \\
         & w-MAPE & 
         0.0418 & 0.0489 & 0.061 & 0.0678 &
         0.0246 & 0.0249 & 0.025 & 0.0257 &
         0.0251 & 0.0274 & 0.0289 & 0.0326 & 0.0355\\ \hline
        \multirow{2}{0.08\textwidth}{DEEPVAR-Proj} & MAPE & 
         0.1621 & 0.1834 & 0.3683 & 0.4697 &
         0.0867 & 0.0839 & 0.0853 & 0.1005 &
         0.138 & 0.1463 & 0.1569 & 0.18 & 0.2211 \\
         & w-MAPE & 
         0.0405 & 0.0477 & 0.0604 & 0.0676 &
         0.0217 & 0.0219 & 0.022 & 0.0227 &
         0.0276 & 0.0299 & 0.0315 & 0.0348 & 0.0378\\ \hline
         \multirow{2}{0.08\textwidth}{NBEATS-BU} & MAPE & 
         0.073 & 0.1287 & 0.276 & 0.3537 &
         0.0254 & 0.0274 & 0.0287 & 0.0458 &
         0.1554 & 0.1621 & 0.166 & 0.173 & 0.2119 \\
         & w-MAPE & 
         0.0183 & 0.0262 & 0.0387 & 0.0458 &
         0.0064 & 0.0067 & 0.0069 & 0.0088 &
         0.0311 & 0.0323 & 0.0338 & 0.0362 & 0.0386\\ \hline
         \multirow{2}{0.08\textwidth}{NBEATS-Proj} & MAPE & 
         0.1167 & 0.1844 & 0.3816 & 0.4895 &
         0.0358 & 0.0351 & 0.0359 & 0.0523&
         0.1131 & 0.1292 & 0.1316 & 0.156 & 0.2299 \\
         & w-MAPE & 
         0.0292 & 0.0346 & 0.0461 & 0.0528 &
         0.0089 & 0.0092 & 0.0093 & 0.0111 &
         0.0226 & 0.0258 & 0.0272 & 0.0316 & 0.0348\\ \hline
         \hline
         \multirow{2}{0.08\textwidth}{\textbf{SLOTH-Opt}} & MAPE & 
         0.0617 & 0.0946 & 0.2274 & 0.2939 &
         \textbf{0.0124} & \textbf{0.0212} & \textbf{0.0226} & \textbf{0.0415}&
         \textbf{0.0323} & \textbf{0.0581} & \textbf{0.0647} & \textbf{0.0877} & 0.1321 \\
         & w-MAPE & 
         0.0154 & 0.0194 & 0.0316 & 0.0368 &
         \textbf{0.0031} & \textbf{0.0041} & \textbf{0.0046} & \textbf{0.0066} &
         \textbf{0.0065} & \textbf{0.0115} & \textbf{0.0134} & 0.0173 & 0.021 \\ \hline
         \multirow{2}{0.08\textwidth}{SLOTH-BU} & MAPE & 
         \textbf{0.0554} & \textbf{0.089} & \textbf{0.2235} & \textbf{0.2916} &
         0.0197 & 0.0287 & 0.0303 & 0.0468&
         0.0345 & 0.0588 & 0.0654 & 0.0883 & 0.1333 \\
         & w-MAPE & 
         \textbf{0.0139} & \textbf{0.0186} & \textbf{0.0306} & \textbf{0.0361} &
         0.0049 & 0.0058 & 0.0062 & 0.0079 &
         0.0069 & 0.0117 & 0.0136 & \textbf{0.0172} & 0.0209\\ \hline
         \multirow{2}{0.08\textwidth}{SLOTH-Proj} & MAPE & 
         0.0682 & 0.0988 & 0.2455 & 0.3118 &
         0.0167 & 0.0255 & 0.027 & 0.0455&
         0.0346 & 0.0586 & 0.0655 & 0.0888 & \textbf{0.1320} \\
         & w-MAPE & 
         0.017 & 0.0209 & 0.0325 & 0.0393 &
         0.0042 & 0.0053 & 0.0056 & 0.0077 &
         0.0069 & 0.0117 & 0.0136 & 0.0172 & \textbf{0.0208}\\ \hline
    \end{tabularx}
    \caption{MAPE and weighted-MAPE(w-MAPE) values across all levels over 5 independent runs. We only show level performances from the part of baselines which perform superior enough. The best performance for each level in three datasets are highlighted.}
    \label{tab:level_results}
    \vspace{-3mm}
\end{table*}

\begin{table}[htbp]
    \centering
    \begin{tabular}{|c|c|c|c|}
        \hline
         \textbf{Deep Model}  & Tourism & Labour  & M5 \\
         \hline
         DeepAR-Proj & 0.1112 & 3.6691 & 15.5745 \\
         DeepVAR-Proj & 0.2431 &  6.4946 &  26.3628\\
         NBEATS-Proj & 1.0372 & 3.6010 & 64.2808 \\
         INFORMER-Proj & 0.6268 & 13.6285  & 53.3686 \\
         DeepAR-BU & 0.2315 & 5.8928  & 24.8748 \\
         NBEATS-BU & 0.9865 & 2.5818  & 11.7834 \\
         INFROMER-BU & 0.3460  & 12.9719 & 53.9309\\
         \hline
         \textbf{SLOTH(Opt)(ours)} & 0.4610 & 6.9624  & 48.2653 \\
          SLOTH(Proj) & 0.1992 & 3.9627 & 18.1219 \\
          SLOTH(BU) & 0.2065 & 4.021  & 19.6753 \\
         \hline
    \end{tabular}
    \caption{Average running time (seconds) over 5 independent runs for baselines such as traditional reconciliation methods and end-to-end methods, as well as our approach. }
    \label{tab:running_time}
\end{table}
As mentioned in section \textbf{Experiment} , the performance difference on w-MAPE and MAPE metrics indicates that the model performs differently across different aggregation levels. 
So we assess the gains in the performance
across all levels presented in Table~\ref{tab:level_results} , which demonstrates that our approach achieves the best performance on both MAPE and weighted-MAPE for 15 out of 25 total levels across three datasets, and our forecasting mechanism, combined with other reconciliation methods, outperforms all other baseline methods across all aggregation levels and bottom levels.  The training running time of deep model for 10 epochs is shown in Table~\ref{tab:running_time}, we don't show the  result of statistic models such as $ARIMA\_BU$  because it is R  scripts which run on CPUs,  it meanless to compare it with deep models running on GPUs.
% The prediction metrics across all levels are presented in Table \ref{tab:level_results}.
% \newcolumntype{t}{>{\hsize=0.5\hsize}c}

\section{Appendix E Ablation Study}

\begin{figure*}
    \centering

    \includegraphics[width=\linewidth]{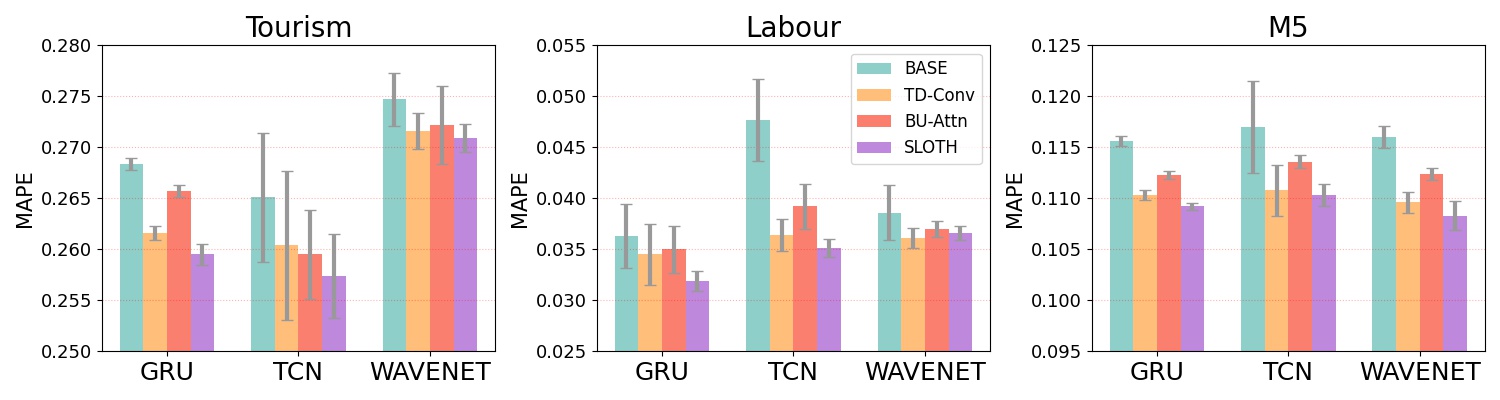}
    \caption{The experiment results of each component in our forecasting over 5 independent runs. We adopt MAPE for evaluation (lower is better). The left figure is the performance on Tourism dataset, the middle figure is for Labour and the right one is for M5.  All figures have three baselines, i.e., GRU, TCN, and WAVENET. The base bars in the figures indicates prediction results only come from the base model without appending any of our top-down convolution or bottom-up attention mechanism, i.e., we only use the feature from GRU, TCN, and WAVENET to produce prediction results. TD-Conv bar means the base model with convolution mechanism, and BU-Attn bar is results of the base model with attention mechanism, and SLOTH bar represents the final results from the combination of both convolution and attention.
    }
    \label{fig:ablation}
    \vspace{-1mm}
\end{figure*}
We conduct ablation studies for each component in our framework, and we demonstrate that we utilize the hierarchical structural information effectively to provide stabler and more accurate results than the state-of-the-art forecasting model. In order to prove that our performance improvement only comes from the forecasting module,
%and validate implicit reconciliation mechanism which prediction result will satisfy aggregation coherency constraints partially, 
we do not perform the explicit reconciliation step after forecasting. 

We first evaluate the validity of the top-down convolution (TD-Conv) mechanism which incorporates the ancestors' information. Our mechanism can be applied to any deep RNN-like model, and we adopt GRU\cite{chung2014empirical}, TCN\cite{bai2018empirical}, and WAVENET\cite{van2016wavenet} as baseline forecasting models to generate dynamic features for each node. As shown in Figure~\ref{fig:ablation}, the orange bar (model with TD-Conv) is always lower than the green bar (base model), therefore, models perform better on the three datasets when the top-down convolution mechanism is applied to integrate dynamic patterns from ancestors. Taking MAPE as the metric, top-down convolution improves the performance by 6.2\% comparing to the baseline. 

\begin{figure}[htbp]
    \centering
    \includegraphics[width=0.4\textwidth]{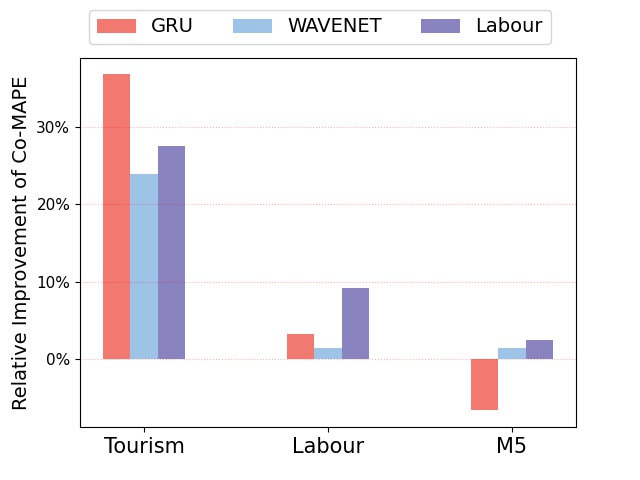}
    \caption{The relative improvement of CO-MAPE of attention mechanism compared against baseline forecasting models}
    \label{fig:co-mape}
    \vspace{-5mm}
\end{figure}

We further compare the performance improvement of the bottom-up attention  mechanism which integrates children's hidden features into the parent node as shown in Figure~\ref{fig:ablation}. We also adopt GRU, TCN, and WAVENET as dynamic feature generators. The results show that the bottom-up attention mechanism helps the forecasting model obtain more informative features and lower the prediction error. This mechanism outperforms the baseline RNN models by 4.3\%.
We also find that the bottom-up attention mechanism improves the coherency without explicit reconciliation as Figure ~\ref{fig:co-mape} demonstrates, wherethe CO-MAPE of the forecasting results from the bottom-up attention integration feature is lower.

\begin{figure}
    \centering
    \includegraphics[width=0.4\textwidth]{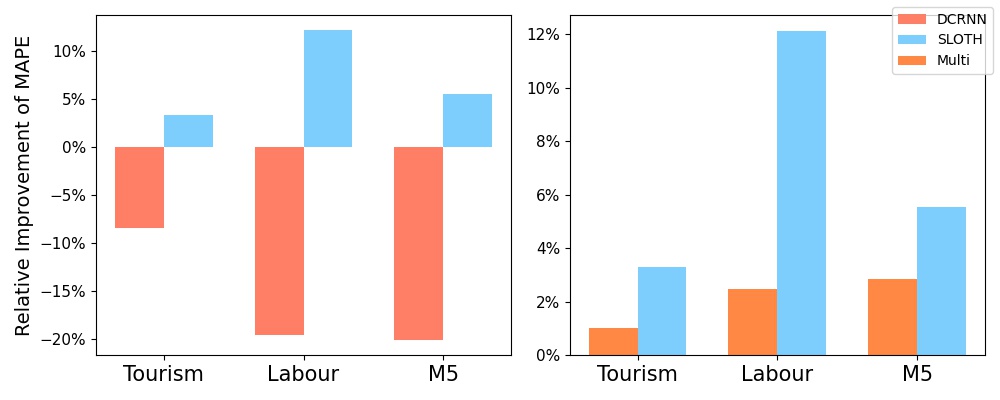}
    \caption{The relative improvement of DCRNN and Multivariate models compared against baseline forecasting models}
    \label{fig:dcrnn_multi}
    \vspace{-3mm}
\end{figure}

As mentioned in \textbf{Method}, the hierarchical structure is a type of the graph structure, but different from the graph time-series scenario because nodes interact with neighbors dynamically; nodes in hierarchical structure only have aggregation relationship.  Nodes at the top levels are virtual or conceptual split points, such as the north-eastern region or a state, and there might exist no physical interaction between nodes at all.
We represent hierarchical structure as a bi-directional adjacent graph where edges only exist between parents and children.  We then adopt the graph time series method, DCRNN, to generate the predictions and compare the performance against baseline models as shown in Figure~\ref{fig:dcrnn_multi}.  We can see DCRNN performs inferior to baseline models, and our explanation is that the graph structure from hierarchical topology is sub-optimal for the representation of the true relationship between nodes, i.e., the mechanism of interaction is inappropriate for the hierarchical structure.  In other words, improper modeling of relationship between nodes induces extra noise for prediction.

HTS forecasting has always been modelled as a multivariate forecasting task in previous methods such as Hier-E2E\cite{rangapuram2021end}, where DeepVar \cite{salinas2019high} is used to extract dynamic patterns and to predict for all nodes, ignoring the hierarchical relationship between each node, but adopts deep neural networks to extract valid similarities that help improve the prediction performance. One limitation of multivariate method is that the increased complexity and the inaccurate relationship from overfitting or underfitting of the deep architecture, causing performance loss for prediction.  We evaluate the performance by taking features of all nodes in the hierarchical structure as input to jointly model the whole system and then generate results for all nodes.  As shown in Figure~\ref{fig:dcrnn_multi}, the multi-variate model performs better than the univariate model, but is still inferior to our proposed approach.

\section{Appendix F M5 Scheduling Task details}
We discuss the details of the designed realistic demand-scheduling task for m5 scheduling task. The penalty setting for each level is shown in Table \ref{tab:penalty_env}.  We would like to minimize the cost of the whole supply chain,  formally, the target is defined as follows:

\begin{table}[htbp]
    \centering
    \begin{tabular}{|c|c|c|c|c|c|}
        \hline
         level& 1& 2& 3& 4 & 5 \\ \hline
         $\mathbf{c}_u$ & 2 & 1 & 0.9 & 0.7 & 0.5 \\ \hline
         $\mathbf{c}_o$ & 50 & 45 & 40 & 35 & 30 \\ \hline
         $\mathbf{q}$ & 0.5 & 0.7 & 0.8 & 1 & 2 \\\hline
    \end{tabular}
    \caption{penalty setting for each level}
    \label{tab:penalty_env}
    \vspace{-3mm}
\end{table}

\begin{equation}
    \label{eq:task_loss}
    \begin{aligned}
        f(\hat{\by}, \by) =\, & \mathbf{c}_u^\mathsf{T} \max(\hat{\by} -\by, \mathbf{0}) + \mathbf{c}_o^\mathsf{T} \max(\by - \hat{\by}, \mathbf{0}) \\
    & + \frac{1}{2}(\hat{\by} - \by)^\mathsf{T} diag(\mathbf{q}) (\hat{\by} - \by),
    \end{aligned}
\end{equation}
where $\hat{\by}$ is the ground truth. In order to simplify computation, we induce $\by_u$ and $\by_o$ to represent the {amount} of underestimation and overestimation, respectively, {Therefore,} our task loss can be reformulated as:
\begin{equation}
    \label{eq:task_cost}
    \begin{aligned}
    &\mathop{\min}_{\by, \mathbf{y}_u, \mathbf{y}_o} \mathbf{c}_u^\mathsf{T} \mathbf{y}_u  + \mathbf{c}_o^\mathsf{T} \mathbf{y}_{o}  - \hat{\by}^\mathsf{T} diag(\mathbf{q})\by + \frac{1}{2}\by ^\mathsf{T} diag(\mathbf{q}) \by ,\\
    &\text{subject to}
    \begin{cases}
    \mathbf{A}\by=\mathbf{0},\\
    \by+\mathbf{y}_u \ge \hat{\by}, \; \mathbf{y}-\mathbf{y}_o\le \hat{\by},\\
    \mathbf{y}, \mathbf{y}_u, \mathbf{y}_o \ge \mathbf{0}.
    \end{cases}
    \end{aligned}
\end{equation}

\section{Appendix G Online Cloud Resource Scheduling Experiments}
\label{sec:alipay_cloud}
\begin{figure}[h]
  \centering
  \includegraphics[width=0.8\linewidth,,height=0.4\linewidth]{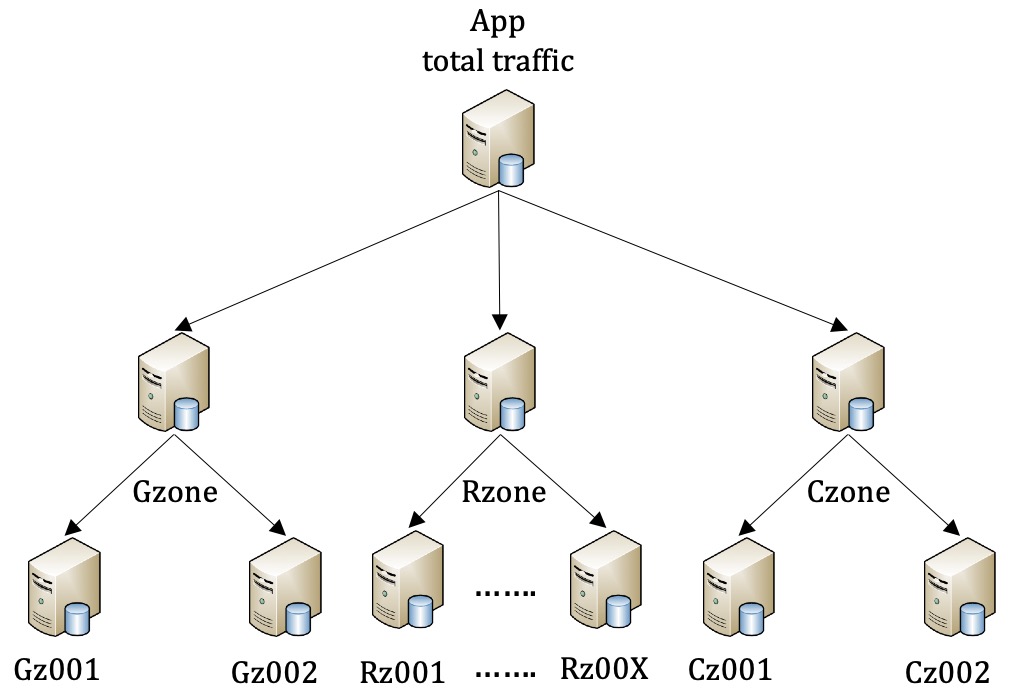}
  \caption{A deployment example of Company's application service.}
  \vspace{-1mm}
  \label{fig:cloud_structure}
\end{figure}

% Online experiments were conducted on the hierarchical sever traffic time series  of Alipay, the world’s leading company of payment technology, to verify the superior performance of our approach. 
Our Company has a giant cluster of application servers to support its complex Internet business, separated into numerous Internet Data Centers (IDCs) \cite{xue_meta_2022,qu_rltpp_2022, xue-2022-hypro}, which are computation facilities that an organization or a service provider relies on. For Company’s IDCs, the deployment of application services presents a hierarchical structure. To be more specific, each deployment includes three levels: the upper layer is called the app deployment unit, and the bottom deployment unit is called the zone, which can be divided into  Gzone, Rzone, and Czone. Each zone is then divided into multiple groups, as shown in Figure~\ref{fig:cloud_structure}.

\begin{figure}
    \centering
    \includegraphics[width=0.5\textwidth]{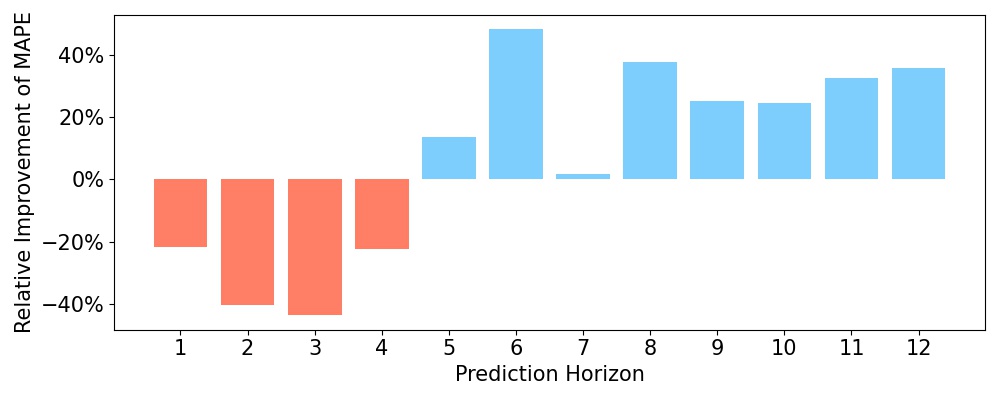}
    \caption{The relative improvement of MAPE compared to the original baseline method over the prediction horizon for the future two hours.}
    \label{fig:cloud}
    \vspace{-3mm}
\end{figure}

We select part of the server hierarchical topology to validate our performance. The hierarchical topology is arranged as follows: 17 IDCs in the bottom level, 3 different zones as the second level, and the root takes the sum of all zones. The frequency granularity is 10min, and we set the prediction length as 12 and the context length as 24.  The experiment results are shown in Figure ~\ref{fig:cloud}, and our approach achieves better performance in a long time horizon forecasting compared to original online baselines.

\section{Appendix H Reproducibility}
\subsection{Detail of SLOTH Architecture}
% \replace{The detail of SLOTH architecture of experiment is as follows, we adopt GRU as RNN component to extract the temporal feature, where the hidden dimension is 128 and layer number is 2;for top down convolution mechanism we take 2-d convolution, the kernel size is (1, index of level num) for each level, then adopts RELU as activation; for bottom up attention mechanism, the hidden dimension of query and key is 128, and the layer number is 1; the dimension of output network is 128, and layer number is 2. For optimization reconciliation, we use default setting of OPTNET.}
{The detail of SLOTH architecture of experiment is as follows, we adopt GRU as RNN component to extract the temporal feature, where the hidden dimension is 128 and layer number is 2;for top down convolution mechanism we take 2-d convolution, the kernel size is (1, index of level num) for each level, then adopts RELU as activation; for bottom up attention mechanism, the hidden dimension of query and key is 128, and the layer number is 1; the dimension of output network is 128, and layer number is 2. For optimization reconciliation, we use default setting of OPTNET.
}

The implementation of pseudocode of SLOTH is shown in \ref{algo_arch}.

\subsection{Hyper Parameters Tuning} 
For all RNN-related neural methods,  the context length of recurrent module is
set in range \{4, 8, 10, 12, 16\} for Tourism dataset, and select from \{4, 8, 12, 16, 24, 32, 40\} for Labour and M5 datasets. For DeepAR and DeepVar methods, the number of rnn layers is selected from range \{1, 2, 3, 4\}, and the number of hidden layers is chosen from range \{32, 64, 128, 256, 512\}.  and lag sequence is set as (1, 4, 8) for Tourism dataset, (1, 2, 3, 4, 5, 6, 7) for Labour and M5 datasets. 
For INFORMER, the layer of encoder
is chosen from range \{ 2, 3, 4\}, the head number of multi-head attention is set as 8, and the dimension of attention is set as 128. For NBEATS, the number of stacks is chosen from range \{32, 48, 64\}, the number of hidden state of fully connect layer is chosen from \{128, 256, 512\}, and the number of blocks is chosen from range \{1, 2, 3\}, and the number of block layers is chosen from range \{2, 4\}. 

For ARIMA and other statistic methods, we follow the setting of HTS package directly.

\subsection{Experiment Setup}
We conduct all our experiments on a server with following configuration, 2 NVIDIA Tesla P100-PCIe GPUs(16GB memory), and 4 Intel(R) Xeon(R) CPU E5-2682 v4 @ 2.50GHz(64 cores) CPUs,  256GB memory and 400TB disk.

\begin{algorithm}[htbp]
\caption{SLOTH Alogorithm}
\label{algo_arch}
\begin{algorithmic}
    \REQUIRE $\mathbf{Y}_{(0,t-1)}, S $
    \STATE $ i \gets 1$, $\bar{\mathbf{H}_t} \gets \mathbf{0}$,
    $\mathbf{H}_t \gets \mathbf{0}$  
    \STATE {/* \textit{temporal feature extraction}} 
    \WHILE{$ i <= n$} 
        \STATE $\bar{H}_t^i = RNN(y_{(0, t-1)}^i, h_0^i) $
        $i \gets i+1$
    \ENDWHILE
    \STATE $ i \gets n$, $\hat{\mathbf{H}}_t \gets \mathbf{0}$,
    \STATE {/* \textit{top down convolution}}
    \WHILE{$ i > 0$} 
        \STATE$\hat{H}_t^i = CNN(\bar{h}^i_t, [\bar{h}^{parents_i}_t]) $
        $i \gets i-1$
    \ENDWHILE
    \STATE {/* \textit{bottom up attention}}
    \STATE $l \gets n_l-1, \Tilde{h}^{\cup_{n_l}}_t = \hat{h}^{\cup_{n_l}}_t, \mathbf{V}_t = 0 $
    \STATE $\mathbf{V}^{n_l}_t = \hat{h}^{\cup_{n_l}}_t,  \cup_i $ \textit{means union of nodes at level i}
    \WHILE{$l > 0$}
        \STATE $\mathbf{Q}_t^{\cup_l} = F(\bar{h}^{\cup_l}_t; \phi_q)$
        \STATE $\mathbf{K}_t^{\cup_{l+1}} = F(\bar{h}^{\cup_{l+1}}_t; \phi_k)$
        \STATE Update $\Tilde{h}_t^{\cup_l}$ according to Equation~\eqref{eq:at-agg}
        \STATE Update $\mathbf{V}_t^{\cup_l} = 
        \Tilde{h}_t^{\cup_l} - \hat{h}^{\cup_{l}}_t + \bar{h}^{\cup_{l}}_t $ 
        % \STATE Update $V_t^{\cup_l}$ based on Equation \ref{eq_value}
        \STATE $l \gets l-1,$
    \ENDWHILE 
    \STATE $ i \gets 1, \mathbf{}{H}_t^i \gets \mathbf{0}, \mathbf{\hat{Y}}_t^i \gets \mathbf{0}$ 
    \STATE {/* \textit{base forecaster}}
    \WHILE{$ i <= n$} 
        \STATE$z_t^i = \sigma(mlp(\tilde{h}_t^i))$
        \STATE $h_{t}^i = (1 -z_t^i) * \tilde{h}_t^i + z_t^i * \bar{h}_t^i $
        \STATE $\hat{y}_t^i = MLP(h_t^i)$
        \STATE $i \gets i+1$
    \ENDWHILE
    \STATE {/* \textit{Optimization Reconciliation}}
    \STATE $\mathbf{Y}_t = OPTNET(\mathbf{\hat{Y}_t, \mathbf{S}})$
    \RETURN $\mathbf{Y}_t$
\end{algorithmic}
\end{algorithm}

\end{document}